\documentclass[iicol,pdflatex,sn-basic]{sn-jnl}%

\usepackage{graphicx}%
\usepackage{multirow}%
\usepackage{amsmath,amssymb,amsfonts}%
\usepackage{amsthm}%
\usepackage{mathrsfs}%
\usepackage[title]{appendix}%
\usepackage{xcolor}%
\usepackage{textcomp}%
\usepackage{manyfoot}%
\usepackage{booktabs}%
\usepackage{algorithm}%
\usepackage{algorithmicx}%
\usepackage{algpseudocode}%
\usepackage{listings}%
\usepackage[numbers]{natbib}
\usepackage{tabularx}
\usepackage{makecell} 
\usepackage{array}
\DeclareUnicodeCharacter{FF1A}{:}

\theoremstyle{thmstyleone}%
\theoremstyle{thmstyletwo}%

\theoremstyle{thmstylethree}%

\begin{document}

\title[Article Title]{TFANet: Three-Stage Image-Text Feature Alignment Network for Robust Referring Image Segmentation}


\author[1]{\fnm{Qianqi} \sur{Lu}}\email{luqianqi@nudt.edu.cn}

\author*[2]{\fnm{Yuxiang} \sur{Xie}}\email{yxxie@nudt.edu.cn}

\author[3]{\fnm{Jing} \sur{Zhang}}\email{m17863974365@163.com}

\author[4]{\fnm{Shiwei} \sur{Zou}}\email{zsw0915@nudt.edu.cn}

\author[5]{\fnm{Yan} \sur{Chen}}\email{2211386457@qq.com}

\author[6]{\fnm{Xidao} \sur{Luan}}\email{xidaoluan@ccsu.cn}

\affil[1,2,3,4,5]{\orgdiv{College of Systems Engineering}, \orgname{National University of Defense Technology}, \orgaddress{ \city{Changsha}, \postcode{410073}, \country{China}}}

\affil[6]{\orgdiv{College of Computer Science and Engineering}, \orgname{Changsha University}, \orgaddress{ \city{Changsha}, \postcode{410000}, \country{China}}}


\abstract{
Referring Image Segmentation (RIS) is a  task that segments image regions based on language expressions, requiring fine-grained alignment between two modalities. However, existing methods often struggle with multimodal misalignment and language semantic loss, especially in complex scenes containing multiple visually similar objects, where uniquely described targets are frequently mislocalized or incompletely segmented.
To tackle these challenges, this paper proposes TFANet, a Three-stage Image-Text Feature Alignment Network that systematically enhances multimodal alignment through a hierarchical framework comprising three stages: Knowledge Plus Stage (KPS), Knowledge Fusion Stage (KFS), and Knowledge Intensification Stage (KIS).
In the first stage, we design the Multiscale Linear Cross-Attention Module (MLAM), which facilitates bidirectional semantic exchange between visual features and  textual representations across multiple scales. This establishes rich and efficient alignment between image regions and different granularities of linguistic descriptions.
Subsequently, the KFS further strengthens feature alignment through the Cross-modal Feature Scanning Module (CFSM), which applies multimodal selective scanning to capture long-range dependencies and construct a unified multimodal representation. This is essential for modeling long-range cross-modal dependencies and enhancing alignment accuracy in complex scenes.
Finally, in the KIS, we propose the Word-level Linguistic Feature-guided Semantic Deepening Module (WFDM) to compensate for semantic degradation introduced in earlier stages. To mitigate language semantic degradation during decoding, WFDM progressively incorporates word-level linguistic cues into the mask generation process, enhancing cross-modal consistency and enabling more precise segmentation, especially when distinguishing between multiple similar objects.
Extensive experiments on the validation subsets of RefCOCO, RefCOCO+, and G-Ref demonstrate that TFANet outperforms state-of-the-art methods, achieving mIoU improvements of 1.84\%, 1.52\%, and 2.29\%, respectively. Ablation and visualization results further confirm that our hierarchical alignment strategy effectively alleviates attention misallocation and semantic loss, enabling more accurate segmentation of uniquely described targets in complex visual scenes.
}

\keywords{Referring Image Segmentation, Multiscale linear Transformer, Cross-Modal Learning, Language Feature Reconstruction}



\maketitle
\clearpage
\section{Introduction}\label{sec1}
Referring Image Segmentation(RIS), a fundamental task at the intersection of computer vision and natural language processing, aims at predicting pixel-level masks of target objects \cite{bib1} in images based on provided natural language descriptions. 
It can be applied in significant practical applications, including language-driven image-based editing \cite{bib2}\cite{bib3}, human-computer interaction \cite{bib4}\cite{bib5}, and autonomous driving assistance systems. 
Unlike traditional single-modal image segmentation tasks, where category conditions are predefined, RIS needs to understand diverse and ambiguous target language expressions to achieve a deeper understanding of images.

\begin{figure}[h]
\centering
\includegraphics[width=\columnwidth]{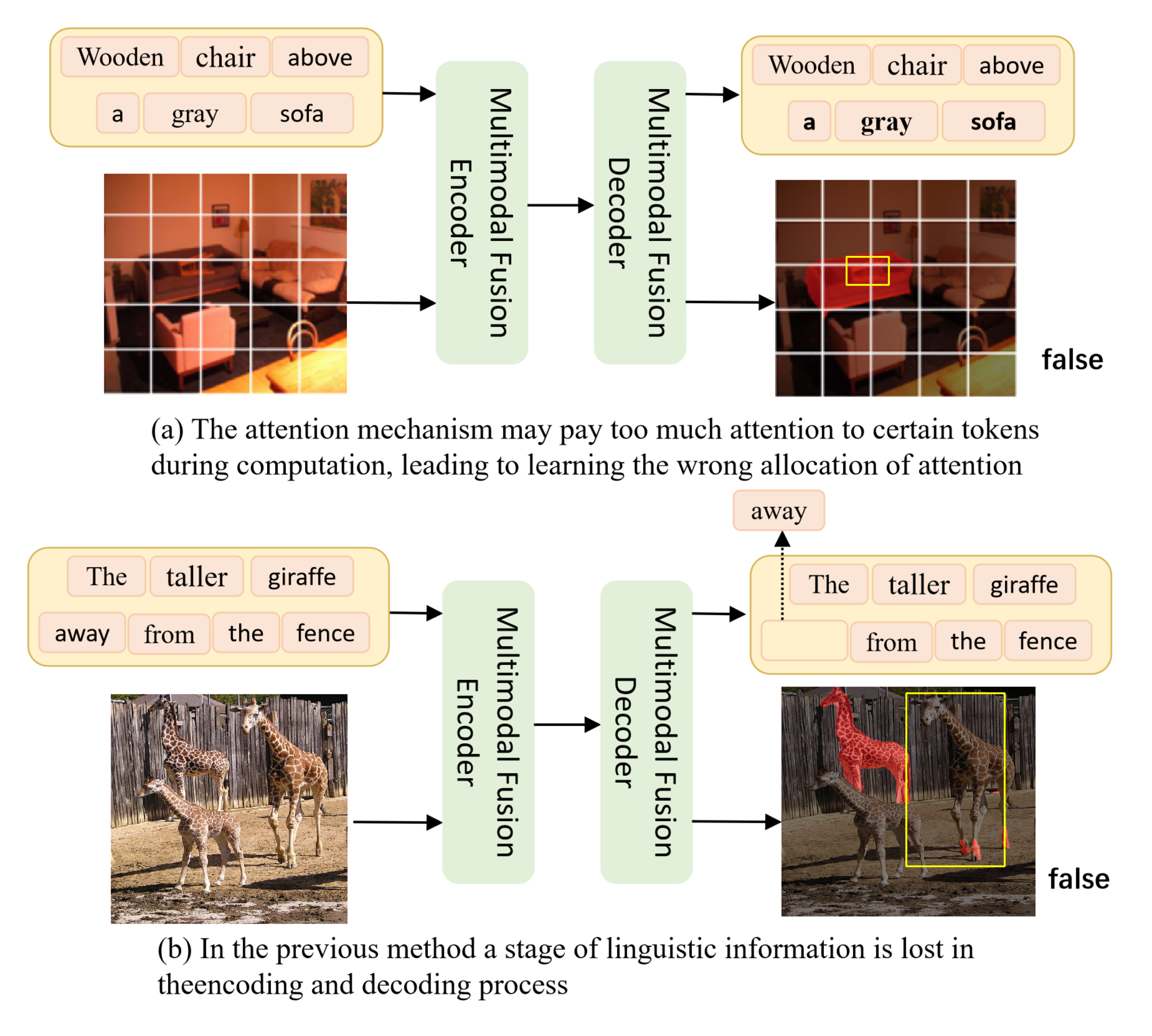} 
\caption{Importance of Multi-stage Feature Alignment in RIS. Many current RIS models exhibit insufficient textual understanding, leading to attention misalignment in complex scenes (Fig.1a). Furthermore, their joint encoding-decoding process can cause semantic loss (Fig.1b), resulting in incorrect segmentation. Red masks denote errors; yellow boxes indicate the correct targets.}
\label{fig1}
\end{figure}

Existing RIS models have explored various strategies to integrate visual and linguistic features. Recently, some approaches have leveraged Attention Mechanisms to enhance cross-modal feature alignment between vision and language. 
For instance, recent studies have introduced Transformer decoders \cite{bib6}\cite{bib7} and Transformer encoder-decoder architectures \cite{bib8}\cite{bib9}\cite{bib10} to achieve cross-modal fusion of visual and linguistic features, leading to substantial improvements in referring image segmentation models. 
However, most existing Attention Mechanisms primarily focus on aligning pixels with word text but fail to fully capture more language semantic and image regional cues from both modalities. These methods fail to establish correspondence between sentence components and image pixels or between image regions and words. This misalignment often leads to incorrect attention allocation, as illustrated in Fig~\ref{fig1}(a), resulting in inaccurate target segmentation. Moreover, previous studies \cite{bib12} have shown that the multimodal alignment encoding and decoding process causes a loss of textual information, as depicted in Fig~\ref{fig1}(b). This loss of linguistic information can further contribute to erroneous segmentation outcomes, potentially leading the model to segment unintended objects.

To address the aforementioned problem, we propose the Three-stage Image-Text Feature Alignment Network (TFANet), which comprises three key stages: Cross-modal Knowledge Plus Stage (KPS), Knowledge Fusion Stage (KFS), and Knowledge Intensification Stage (KIS). 
In KPS, our objective is to enhance multiscale interactions and strengthen the complementarity between abstract textual knowledge and concrete visual knowledge. 
To achieve this, we design the Multiscale Linear Cross-Attention Module (MLAM), which facilitates bidirectional multiscale knowledge interaction between word and sentence-level textual knowledge and pixel and region-level visual knowledge. 
This mechanism enables a more comprehensive understanding of multimodal features and improves attention allocation accuracy. Simultaneously, the time complexity of this component is reduced to linear by optimizing the attention mechanism’s computational method, ensuring that the overall model maintains linear time complexity. 
In the KFS, TFANet focuses on enhancing the global coherence and alignment accuracy between textual and visual modalities. Rather than merely aggregating features, this stage introduces the the Cross-modal Feature Scanning Module (CFSM) that selectively emphasizes semantically relevant patterns across both spatial and channel dimensions. 
By capturing long-range dependencies and dynamically adjusting to multimodal context, KFS effectively transforms loosely correlated features into a unified and discriminative joint representation, enabling the precise mapping of fine-grained verbal cues to their corresponding visual regions.
Moreover, to address the persistent challenge of the loss of linguistic knowledge in the encoding process and the insufficient text-image interaction in the decoding stage, we introduce the Knowledge Intensification Stage (KIS). Within this stage, we design a Word-level Linguistic Feature-guided Semantic Deepening Module (WFDM) to refine feature fusion by continuously injecting word-level linguistic information during mask generation through gradual iterations. By establishing direct guidance from target representations to mask formation and complementing by linguistic information, this architecture produces refined segmentation masks.

To evaluate the effectiveness of TFANet, we conducted extensive experiments on three standard benchmark datasets: RefCOCO \cite{bib13}, RefCOCO+ \cite{bib13}, and G-Ref \cite{bib14},  \cite{bib15}. The proposed three-stage feature alignment framework consistently outperforms state-of-the-art methods, achieving mIoU gains of 1.84\%, 1.52\%, and 2.29\% on the validation subsets of RefCOCO, RefCOCO+, and G-Ref, respectively.
 
Overall, the main contributions of this paper are summarized as follows:

\begin{itemize}
\item We propose TFANet, a novel framework for referring image segmentation. This framework systematically enhances text-image alignment through a three-stage cross-modal feature alignment pipeline (KPS, KFS, KIS) specifically designed to address challenges posed by long text and complex scenarios.

\item To address the challenges of attention misallocation and semantic loss in referring image segmentation, we design three task-specific modules: MLAM, CFSM and WFDM. Each module targets a key stage in the segmentation pipeline—enabling accurate multiscale alignment, global context integration, and language semantic reconstruction, respectively. Together, these modules enhance the model’s ability to identify and segment uniquely described targets with improved precision and robustness.

\item Our TFANet establishes a new benchmark in referring image segmentation, achieving state-of-the-art results on RefCOCO, RefCOCO+, and G-Ref datasets. Extensive experiments not only validate its superior performance but also highlight its robustness in handling complex linguistic and visual interactions.

\end{itemize}

\section{Related Works}\label{sec2}

\subsection{Referring Image Segmentation}\label{subsec2}

Referring Image Segmentation (RIS) aims to precisely delineate target regions in images based on natural language descriptions. Early RIS architectures \cite{bib16}\cite{bib17}\cite{bib18} primarily relied on concatenation-based fusion, where visual and linguistic features were first concatenated and then processed through convolutional neural networks (CNNs).  
Subsequent advancements introduced Recurrent Neural Networks (RNNs) and dynamic convolution mechanisms \cite{bib19}, progressively refining segmentation masks and improving overall task performance. 
Building on this, several studies \cite{bib10,bib20} explored the timing of multimodal fusion and highlighted the superiority of early fusion strategies, which led to architectures that enforce visual-linguistic alignment in the initial encoding stages. This shift paved the way for Transformer-based models that could jointly model long-range dependencies in both modalities. Representative examples include MDETR \cite{bib7} and VLT \cite{bib6}, which utilized Transformer decoders for cross-modal fusion. LAVT \cite{bib10} went further by integrating vision-language fusion into a hierarchical Swin Transformer \cite{bib25} backbone. In a similar vein, ReSTR \cite{bib8} and CRIS \cite{bib26} employed dual-stream Transformer encoders and multimodal decoder structures to enhance alignment.

More recent works diversified the formulation of RIS. PolyFormer \cite{bib27} and SeqTr \cite{bib28} modeled segmentation as contour prediction using sequential point outputs, while GRES \cite{bib29} and CGFormer \cite{bib30} approached it as proposal-level classification, interpreting Transformer queries as object regions. These methods reflect a broader trend toward exploiting Transformer flexibility for RIS, though they often come with higher computational costs.
To tackle this, recent studies have shifted attention toward lightweight and efficient RIS models. For instance, CrossVLT \cite{bib62} introduced cross-layer interaction mechanisms for more adaptive query-region alignment. PTQ4RIS \cite{bib63} proposed a quantization pipeline tailored for RIS, achieving compression without sacrificing accuracy. DETRIS \cite{bib61} advanced the field by employing parameter-efficient tuning (PET) with low-rank propagation and text adapters, significantly reducing model updates while maintaining strong performance. These efforts collectively point toward RIS models that are not only accurate but also scalable and hardware-friendly.

In contrast to existing approaches, ReMamber \cite{bib11} introduces a novel multimodal architecture based on Mamba \cite{bib31} , integrating State Space Models (SSM) and Cross-Scanning Modules (CSM) to scan hybrid features across channel and spatial dimensions. This design facilitates efficient cross-modal fusion in visual-language tasks. 

However, several challenges persist, including insufficient modeling of deep intermodal relationships and imprecise boundary delineation. Existing architectures struggle to establish effective interactions between language and vision, and may result in the loss of linguistic knowledge during feature fusion. To address these limitations, TFANet introduces a three-stage modal alignment framework that systematically enhances cross-modal knowledge integration through task-specific interaction mechanisms. This approach ultimately produces accurate segmentation masks.

\subsection{Transformer for Cross-Modal Alignment}\label{subsec2}

Originally developed for sequence modeling in Natural Language Processing (NLP) tasks \cite{bib32}, the Transformer architecture has rapidly transformed the field due to its powerful global attention capabilities \cite{bib33,bib34,bib35}. Its success soon extended to Computer Vision (CV) domains, achieving state-of-the-art results in image classification, action recognition, and related tasks \cite{bib36,bib37,bib38,bib39,bib40}. With the emergence of cross-modal learning, research has increasingly focused on adapting Transformers for visual-linguistic integration. Prominent examples include CLIP \cite{bib41}, which utilizes dual encoders and contrastive learning, and UniT \cite{bib42}, which unifies multi-task cross-modal learning through a shared Transformer backbone.

In the context of Referring Image Segmentation (RIS), Transformer-based models have proven particularly effective for aligning visual and textual semantics. Prior works \cite{bib10,bib27,bib43,bib44} demonstrate that Transformers excel at propagating fine-grained language cues to guide visual decoding. However, traditional token-wise attention often results in over-concentration on a few salient tokens, leading to suboptimal localization—especially in scenes with multiple similar objects. This bottleneck limits the model’s ability to maintain consistent and context-aware alignment between textual descriptors and visual targets.

To overcome these limitations, TFANet proposes the MLAM, a novel bidirectional cross-attention mechanism. By implementing multiscale linear attention computations, MLAM facilitates hierarchical cross-modal interactions while maintaining linear computational complexity, thereby achieving enhanced multimodal fusion efficiency.

\section{Methods}\label{sec3}

\subsection{Overview}\label{subsec2}


\begin{figure*}[htbp] 
\centering
\includegraphics[width=0.9\textwidth]{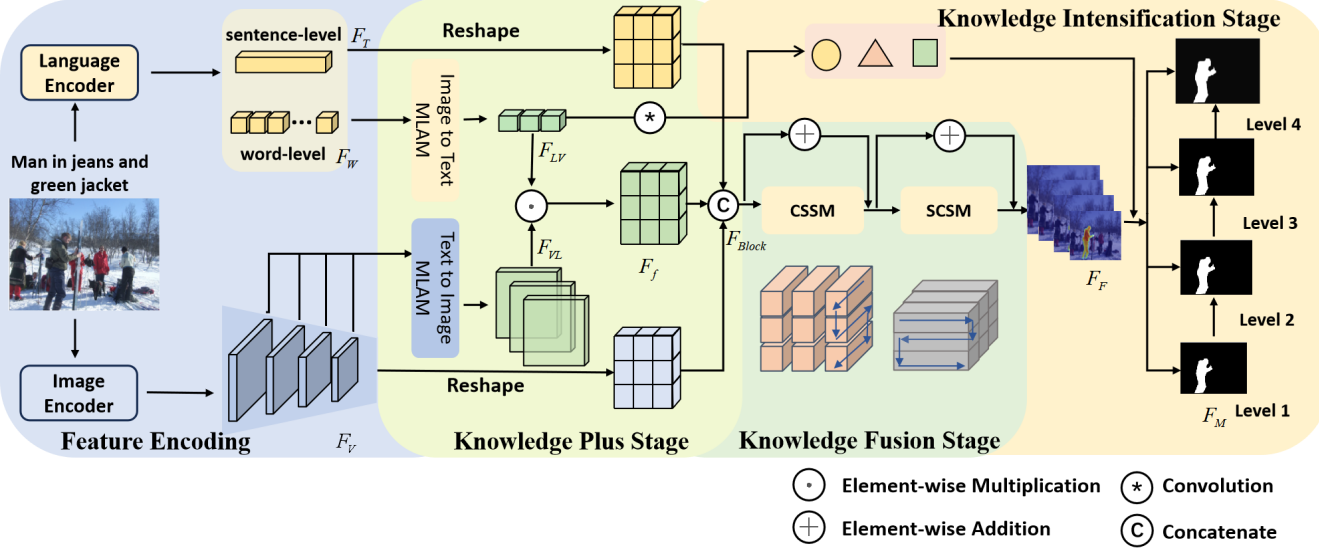} 
\caption{ Overview of the proposed TFANet. The architecture features a hierarchical structure with three core stages: KPS, KFS, KIS. TFANet incorporates three innovative components corresponding to each stage: the MLAM for cross-scale feature correlation, the CFSM for multimodal pattern discovery, and the WFDM for context-aware refinement. }
\label{fig2}
\end{figure*}

In this paper, we propose a Three-stage Feature Alignment Network (TFANet) for robust referring image segmentation as illustrated in Fig~\ref{fig2}. The framework adopts a hierarchical architecture consisting of three core processing stages: Knowledge Plus Stage (KPS), Knowledge Fusion Stage (KFS), and Knowledge Intensification Stage (KIS). Specifically, this paper designs three novel modules to facilitate cross-modal interactions at different stages: the Multiscale Linear Cross-Attention Module (MLAM), the Cross-Modal Feature Scanning Module (CFSM), and the Word-level Linguistic Feature-guided Semantic Deepening Module (WFDM). By progressively refining visual-linguistic feature integration, TFANet achieves multi-scale semantic alignment while enhancing segmentation precision through complementary feature fusion.

In particular, the input image and referring expression are first extracted by VMamba \cite{bib45} and CLIP \cite{bib41} to obtain image features $F_{V}$, sentence-level text features $F_{T}$, and word-level text features $F_{W}$, respectively. After obtaining $F_{V}$, it is fed into the KPS along with $F_{W}$. During this stage, the MLAM facilitates multi-scale bidirectional interactions between the image and text features to generate linguistic features fused with visual information $F_{LV}$ and visual features  which is fused with linguistic information $F_{VL}$. By multiplying these two, the fusion feature $F_{f}$ is obtained, which is then concatenated with both $F_{T}$ and $F_{V}$ to form the final fusion feature block $F_{Block}$. 
Subsequently, the fused feature blockis input into the KFS for further feature fusion. The CFSM then scans the $F_{Block}$ along both the channel and spatial dimensions, enabling long-range selective interactions between features and producing the multi-modal feature $F_{F}$. Finally, the $F_{F}$ are input into the KIS, where the WFDM continuously injects linguistic information to enhance the cross-modal  knowledge interaction during the mask generation process, resulting in the final mask feature map. In summary, the KPS facilitates multi-scale interactions between textual and image features, while the KFS enables deeper cross-modal feature integration. Furthermore, the KIS refines word-level linguistic information during the mask generation stage, guiding visual features to emphasize the target and assisting the model in accurately locating and segmenting the target object.

\subsection{Image and Text Feature Encoding}\label{subsec2}

\bmhead {Text Encoder} For a given linguistic expression $L\in R^{L}$ , this paper employs CLIP \cite{bib41} to extract the word-level text features $F_{W} \in R^{L\times C}$ and the sentence-level text features $F_{T}\in R^{1 \times C}$ . where , C represents the feature dimension and L denotes the length of the description.

\bmhead {Visual Encoder} Previous RIS studies, such as LAVT \cite{bib10}, have demonstrated that early-stage multimodal feature fusion within the visual encoding process improves segmentation performance. Meanwhile, ReMamber \cite{bib11} introduced the innovative use of VMamba as an image encoder, significantly enhancing both encoding speed and performance. Building on these advancements, this paper proposes an image encoder that effectively integrates both visual and linguistic features to further enhance cross-modal feature representation. For a given image $F_{I} \in R^{H\times W\times 3}$ , we process it using the VMamba \cite{bib45} encoder, which consists of four stages and generates the corresponding feature representations $\left\{F_{Vi}\right\}_{i=1}^{4}\in\mathbb{R}^{H_{i}\times W_{i}\times C_{i}}$ , where, $H_{i}$ , $W_{i}$ and $C_{i}$ denote the height, width, and number of channels of the image, respectively. The alignment of linguistic features is realized through the introduction of our TFANet, which operates between the stages of visual coding processing. Specifically, we denote TFANet as $\{\delta_ i\mid i\in\{1,2,3,4\}\}$, and the four stages in VMamba as ${F_{V_i} \mid i \in \{1,2,3,4\}}$. The multi-modal computational procedure for the encoding phase can be formed as follows:

\begin{equation}F_{V_1}=VSS_{1}(I)\end{equation}
\begin{equation}F_{i}=\delta_{i}\left(F_{V_i},F_{W}\right),i\in\{1,2,3,4\}\end{equation}
\begin{equation}F_{V_i}=VSS_{i}\left(F_{V_i-1}+\sigma(F_{i})\right),i\in\{2,3,4\}
\end{equation}

The above $\{F_{i}\}_{i=2}^{4}\in\mathbb{R}^{H_{i}\times W_{i}\times C_{i}}$ represents the features after cross-modal alignment and $\sigma$ indicates the downsampling operation. In stages of the visual encoder beyond the first stage, we employ TFANet to fuse visual and linguistic features, generating multimodal features that are then used as input for the subsequent stages. Finally, the multi-modal features from all three stages are obtained. 

\subsection{Knowledge Plus Stage}\label{subsec3}
As the first stage of TFANet, the Knowledge Plus Stage (KPS) is designed to facilitate the interaction between the abstract knowledge in the text and the concrete knowledge in the image. This interaction generates a linguistic featurfused with visual information $F_{LV}$ and a visual feature fused with linguistic information $F_{VL}$, achieved through a Multiscale Linear bidirectional Cross-Attention Module that performs multiscale cross-attention computations. By multiplying the above two features, the two modalities are merged into a single feature, which is then spliced with the other components to finally form the fusion feature block $F_{Block}$. Specifically, KPS consists of two main parts: multi-scale linear bidirectional attention between image and text, and the construction of the fusion cube.

\bmhead {Multiscale Linear Cross-Attention Module} To distinguish the target object from the background, visual and linguistic representations must be unified across modalities. A common approach is to pair each pixel token with its corresponding lexical token, enabling the learning of a multimodal representation capable of differentiating between the "target" and "background" classes. Previous studies have proposed various mechanisms to address this challenge, including dynamic convolution \cite{bib46}, connectivity mechanisms \cite{bib16}\cite{bib18}\cite{bib46} cross-modal attention \cite{bib20}\cite{bib48}\cite{bib49}\cite{bib50}\cite{bib51} and graph neural networks \cite{bib52}. Notably, in addition to single-point representations, both local visual regions and text sequences store essential information for a comprehensive understanding of multimodal features. Consequently unlike most existing cross-modal attention mechanisms, our proposed Multiscale Linear Bidirectional Cross-Attention Module (MLAM) extends the conventional pixel-to-word computation to multiple scales, including pixel -to-word, region-to-word, pixel-to-phrase, and region- to-phrase interactions at different scales, leveraging a multiscale computational approach. Furthermore, inspired by X-VIT \cite{bib53}, we enhance traditional attention computation efficiency by replacing the Softmax operation with X-Norm and employing the multiplicative union law, thereby reducing the computational complexity from quadratic to linear.

\begin{figure}[h]
\centering
\includegraphics[width=\columnwidth]{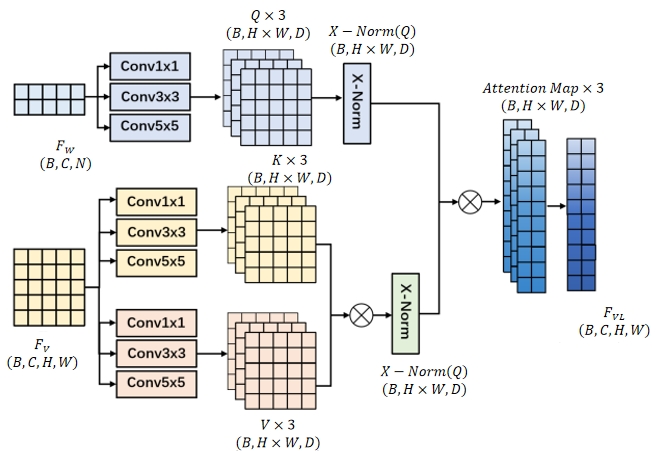} 
\caption{The flow of Text-to-Image Alignment for the Multiscale Linear Cross-Attention Module. First, a set of queries is generated from the input visual feature map $F_{W}$ using convolutional kernels of varying sizes. Similarly, a corresponding set of keys and values is derived from the input language feature  map $F_{V}$ through the same operation. Then, the transpose of each key is multiplied by its corresponding value, obtained using the same convolutional kernel size. The X-Norm operation is then applied, followed by multipli-cation with the query, which has also undergone X-Norm processing. This results in a set of feature maps, which are subsequently weighted to obtain the final $F_{VL}$.}
\label{fig3}
\end{figure}

\bmhead{Text-to-Image Alignment} The schematic diagram of MLAM is shown in Fig~\ref{fig3}. When the input visual and textual features are defined, MLAM enhances multimodal features across multiple scales in the following step. For clarity, the subscripts of the features at each stage are omitted. To compute the language-to-image attention, the word-level linguistic features $F_{W}$ and the encoded image features $F_{V}$ are first processed. Specifically, the word-level linguistic features are transformed into query Q while the visual features are converted into key K and value V through a set of embedding matrices $\left\{W_{q_{i}}\right\}_{i=1}^{3}$, $\left\{W_{k_{i}}\right\}_{i=1}^{3}$, $\left\{W_{v_{i}}\right\}_{i=1}^{3}$ as shown follows:

\begin{equation}\{Q_{i}\}_{i=1}^{3}=\{W_{q_{i}}\times F_{W}\}_{i=1}^{3}\end{equation}

\begin{equation}\{K_{i}\}_{i=1}^{3}=\{W_{k_{i}}\times F_{V}\}_{i=1}^{3}\end{equation}

\begin{equation}\{V_{i}\}_{i=1}^{3}=\{W_{v_{i}}\times F_{V}\}_{i=1}^{3}\end{equation}

It is important to note that the embedding matrix is spatially realized using 1×1, 3×3 and 5×5 convolutions. The embedding matrix enables to obtain a set of query $\{Q_{i}\}_{i=1}^{3}$, key $\{K_{i}\}_{i=1}^{3}$, value$\{V_{i}\}_{i=1}^{3}$. The second step of attention computation involves the queries, keys, and values, which are derived from convolutions with kernels of the same size. The traditional cross-attention is then computed as follows: the similarity or correlation score between the query and the key are calculated using a dot product operation. The Softmax operation normalizes the similarity score of each row. Finally, it is multiplied with the value of value to get the final attention feature map.

\begin{equation}
A(\mathbf{x})=\mathrm{Softmax}(\frac{QK^{T}}{\sqrt{d_{k}}})V
\end{equation}

The attention computation proposed in this paper differs from the traditional attention mechanism in two key aspects. First, we replace the Softmax function with the X-Norm operation to introduce a nonlinear weighting mechanism, which enhances attention focus on relevant information and stabilizes the training process. Second, the attention computation in this paper transforms $QK^{T}V$ into $Q(K^{T}V)$ using the multiplicative union law, where the transpose of the key is first computed and multiplied by the value, followed by the X-Norm operation applied separately to both $(K^{T}V)$ and the query, then the two results are multiplied. The computational complexity of each matrix multiplication is reduced from $O(N^{2})$ in the traditional approach to $O(N)$, significantly improving computational efficiency. The specific calculation formula is as follows：

\begin{equation}A_{VL}(x_i)=\mathrm{X}_{Nom}(Q_{i})(\mathrm{X}_{Nom}(K_{i}^{T}V_{i}))\end{equation}
\begin{equation}X_{_{Norm}}(x)=\frac{\gamma x}{\sqrt{\sum_{i=0}^{h}||x||^{2}}}\end{equation}

Where $\gamma$ is a learnable parameter and h represents the number of embedding dimensions. $\left\|\cdot\right\|$ corresponds to the standard $L_2Norm$ , which is applied to the patches of $(K^{T}V)$  and the filters of Q. The final step involves performing a weighted summation of the multiscale attention maps to obtain the linguistic feature map $F_{VL}$, which incorporates visual information.

\begin{equation}F_{VL}=\sum_{i=1}^{3}w_{i}\times A_{VL}(x_i)\end{equation}

\bmhead{Image-to-Text Alignment}Similar to the Text-to-Image Alignment process described earlier, the alignment of multi-scale multimodal features is conducted in three steps, utilizing the input visual and textual features. In the first step, image features are transformed into query Q, while the word-level linguistic features are converted into key K and value V through a set of embedding matrices $\left\{W_{q_{i}}\right\}_{i=1}^{3}$, $\left\{W_{k_{i}}\right\}_{i=1}^{3}$, $\left\{W_{v_{i}}\right\}_{i=1}^{3}$. Image and text features are then processed with convolution kernels of sizes 1×1, 3×3, and 5×5, generating the corresponding query $\{Q_{i}\}_{i=1}^{3}$, key $\{K_{i}\}_{i=1}^{3}$, value$\{V_{i}\}_{i=1}^{3}$.

\begin{equation}\{Q_{i}\}_{i=1}^{3}=\{W_{q_{i}}\times F_{V}\}_{i=1}^{3}\end{equation}

\begin{equation}\{K_{i}\}_{i=1}^{3}=\{W_{k_{i}}\times F_{W}\}_{i=1}^{3}\end{equation}

\begin{equation}\{V_{i}\}_{i=1}^{3}=\{W_{v_{i}}\times F_{W}\}_{i=1}^{3}\end{equation}

Subsequently, attention is calculated in the same way as above.
\begin{equation}A_{LV}(x_i)=\mathrm{X}_{Norm}(Q_{i})(\mathrm{X}_{Norm}(K_{i}{}^{T}V_{i}))\end{equation}

Finally, a weighted summation operation is applied to the multi-scale attention map to obtain the visual feature map $F_{LV}$ which incorporates linguistic information.

\begin{equation}F_{LV}=\sum_{i=1}^{3}w_{i}\times A_{LV}(x_i)\end{equation}

\bmhead{Construction of fusion feature blocks}After performing the aforementioned multiscale linear bidirectional cross-attention operations, we obtain text features with enhanced image information $F_{VL}$ and image features with enhanced text information$F_{LV}$. However, relying solely on these multiscale attention mechanisms may not suffice to capture the intricate relationships between different modalities. To address this, we further align the modality-specific features into a unified representation space by performing matrix multiplication between $F_{VL}$ and $F_{LV}$ to compute local interactions, which results in the fusion feature $F_{f}$. 

\begin{equation}F_f=F_{VL}W_V\cdot(F_{LV}W_L)^T\in\mathbb{R}^{H\times W\times L}\end{equation}

To improve the processing of high-dimensional features, we apply a single convolutional layer to transform $F_{f}$ into $F_c\in\mathbb{R}^{H\times W\times C}$ and scale them to match the size of the image features $\tilde{F}_{LV}\in\mathbb{R}^{H\times W\times C}$.

\begin{equation}\tilde{F}_{VL}=E(F_{VL})\in\mathbb{R}^{H\times W\times C}\end{equation}
\begin{equation}\tilde{F}_{LV}=E(F_{LV})\in\mathbb{R}^{H\times W\times C}\end{equation}

Where, $E(\bullet)$ denotes the operation of expanding the input tensor to align with the dimensions of the image features. Subsequently, we concatenate $F_{c}$, $\tilde{F}_{VL}$ and $\tilde{F}_{LV}$ along the channel dimension to form the fusion feature block ${F}_{Block}\in\mathbb{R}^{H\times W\times(C_{i}+C_{i}+C_{c})}$.

\begin{equation}F_{Block}=[\tilde{F}_{VL},\tilde{F}_{LV},F_{c}]\end{equation}

\subsection{Knowledge Fusion Stage}\label{subsec4}
Knowledge Fusion Stage (KFS) is designed to facilitate further fusion of concrete knowledge features in vision and abstract knowledge of linguistic features over long distances. Analyzing each spatial region of the feature map from a purely local perspective fails to account for the essential interactions with distant spatial regions within the global context, which are crucial for semantic understanding and segmentation tasks. 
To address this challenge, the visual-linguistic fusion feature block ${F}_{Block}$ , obtained in the previous stage, serves as input for the Cross-Modal Feature Scanning Module (CFSM). This module scans ${F}_{Block}$, across both channel and spatial dimensions, capturing long-range dependencies between the linguistic and visual features. This process enables selective long-range interactions, ultimately producing a robust multimodal feature representation for the task, denoted as ${F}_{F}$. 

Specifically, to capture long-range dependencies between words and pixels, the CFSM first employs the Channel Selective Scanning Model (CSSM) to perform one-dimensional channel selective scanning along the channel dimension. The spliced hybrid feature block is treated as an ordered sequence along the channel dimension, enabling cross-channel communication to facilitate visual-linguistic fusion. This results in an intermediate feature, which is then combined with the original ${F}_{Block}$ to obtain $F_{Channel}$. Next, the Spatial Cross-modal Scanning Model (SCSM) performs 2D spatial selective scanning on each feature map to learn cross-modal communication independently within each channel of the intermediate feature.  The selective scanning mechanism is employed to establish long-range dependencies between words and spatial regions, ultimately producing the feature maps ${F}_{Space}$ .The final enhanced cross-modal feature map ${F}_{F}$ is obtained by summing ${F}_{Space}$ with the feature map ${F}_{Channel}$, and then forwarded to the next stage for generating the final mask.

\begin{equation}F_{\mathrm{Channel}}=CSSM\left(F_{Block}\right)\end{equation}
\begin{equation}F_{space}=SCSM\left(F_{Block}+F_{c}\right)\end{equation}
\begin{equation}F_{F}=F_{S\mathrm{pace}}+F_{\mathrm{Charmel}}\end{equation}

\subsection{Knowledge Intensification Stage}\label{subsec5}
Specifically, as illustrated in Fig~\ref{fig2}, we first generate a dynamic kernel $K_{i}\in\mathbb{R}^{B\times C\times N}$ , for the linguistic information ${F}_{LV}$ by performing a one-dimensional convolution on the word-level linguistic features derived from the first stage at each previous level. The output ${K}_{i}$ of the dynamic kernel is then up-sampled using bilinear interpolation to match the size of the current level in the hybrid feature block,${K}_{i}\in\mathbb{R}^{B\times C\times H\times W}$.

\begin{equation}K_{i}=\omega(\delta(F_{LV_{i}})),i\in\{1,2,3,4\}\end{equation}

Where $\delta(\bullet)$ denotes a one-dimensional 1×1 convolution operation, $\omega(\bullet)$ represents a bilinear interpolation upsampling operation, i corresponds to the different encoding stages of VMamba, and C, N indicate the number of channels and the dimensions of the final convolution kernel, respectively. Then, the dynamic kernel and the different encoding stages are element-wise multiplied with the enhanced hybrid features $\{F_{F1},F_{F2},F_{F3},F_{F4}\}$ which  is  formed  in  the  first  two stages to obtain the Mask maps at different levels $\{F_{M1},F_{M2},F_{M3},F_{M4}\}\in\mathbb{R}^{B\times C\times H\times W}$.

\begin{equation}F_{Mi}=K_{i}\times F_{Fi},i\in\{1,2,3,4\}\end{equation}

The masks at different levels are aggregated through upsampling and weighted summation to generate a final aggregated mask.

\begin{equation}M_{1}=F_{M1}\end{equation}
\begin{equation}M_{i+1}=\alpha\mathrm{Up}\left(M_{i}\right)+(1-\alpha)F_{Mi+1},i\in\{1,2,3\}\end{equation}

Where $\mathrm{Up(\bullet)}$ denotes the bilinear upsampling operation and $\alpha$ controls the fusion ratio between strata. The final feature mapis projected into the two stratum score maps using a 1×1 convolution.

\begin{table*}[ht]
\centering
\caption{Comparisons with the state-of-the-art approaches on three benchmarks.}
\renewcommand{\arraystretch}{0.75}  
\setlength{\tabcolsep}{7pt}         
\footnotesize
\begin{tabular}{llcccccccc}
\toprule
\multirow{2}{*}{Metric} & \multirow{2}{*}{Model)} & \multicolumn{3}{c}{RefCOCO} & \multicolumn{3}{c}{RefCOCO+} & \multicolumn{2}{c}{G-Ref} \\
\cmidrule(lr){3-5} \cmidrule(lr){6-8} \cmidrule(lr){9-10}
 & & val & testA & testB & val & testA & testB & valU & testU \\
\midrule
\multirow{12}{*}{mIoU}
 & MCN~\cite{bib47} (CVPR-20) & 62.44 & 64.20 & 59.71 & 50.62 & 54.99 & 44.69 & 49.22 & 49.40 \\
 & EFNet~\cite{bib20} (CVPR-21) & 62.76 & 65.69 & 59.67 & 51.50 & 55.24 & 43.01 & 51.90 & - \\
 & LTS~\cite{bib55} (CVPR-21) & 65.43 & 67.76 & 63.08 & 54.21 & 58.32 & 48.02 & 54.40 & 54.25 \\
 & VLT~\cite{bib6} (ICCV-21) & 65.65 & 68.29 & 62.73 & 55.50 & 59.20 & 49.36 & 52.99 & 56.65 \\
 & CRIS~\cite{bib26} (CVPR-22) & 70.47 & 73.18 & 66.10 & 62.27 & 68.08 & 53.68 & 59.87 & 60.36 \\
 & ReSTR~\cite{bib8} (CVPR-22) & 67.22 & 69.30 & 64.45 & 55.78 & 60.44 & 48.27 & 54.48 & - \\
 & ReLA~\cite{bib29} (CVPR-23) & 73.82 & 76.48 & 70.18 & 66.04 & 71.02 & 57.65 & 65.00 & 65.97 \\
 & PTQ4RIS~\cite{bib63} (ICRA-25) & 74.31 & 76.63 & 70.61 & 66.69 & 71.47 & 60.01 & 65.91 & 66.01 \\
 & CRFormer~\cite{bib56} (ICMR-24) & 75.26 & 77.38 & 71.92 & 66.98 & 71.74 & 59.32 & 65.97 & 66.86 \\
 & DETRIS~\cite{bib61} (AAAI-25) & 76.00 & 78.20 & 73.50 & 68.90 & 74.00 & 61.50 & 67.90 & 68.10 \\
 \midrule
 & \textbf{TFANet (Ours)} & \textbf{77.84} & \textbf{79.42} & \textbf{75.16} & \textbf{70.42} & \textbf{74.05} & \textbf{62.95} & \textbf{70.19} & \textbf{72.10} \\
\midrule
\multirow{13}{*}{oIoU}
 & CMCP~\cite{bib21} (CVPR-20) & 61.36 & 64.53 & 59.64 & 49.56 & 53.44 & 43.23 & - & - \\
 & LSCM~\cite{bib22} (ECCV-20) & 61.47 & 64.99 & 59.55 & 49.34 & 53.12 & 43.50 & - & - \\
 & EFNet~\cite{bib20} (CVPR-21) & 62.76 & 65.69 & 59.67 & 51.50 & 55.24 & 43.01 & 51.93 & - \\
 & BUSNet~\cite{bib57} (CVPR-21) & 63.27 & 66.41 & 61.39 & 51.76 & 56.87 & 44.13 & - & - \\
 & LAVT~\cite{bib10} (CVPR-22) & 72.73 & 75.82 & 68.79 & 62.14 & 68.38 & 55.10 & 61.24 & 62.09 \\
 & CrossVLT~\cite{bib62} (TMM-24) & 73.44 & 76.16 & 70.15 & 63.60 & 69.10 & 55.23 & 62.68 & 63.75 \\
 & CoupAlign~\cite{bib43} (NeurIPS-22) & 74.70 & 77.76 & 70.58 & 62.92 & 68.34 & 56.69 & 62.84 & 62.22 \\
 & DDMI~\cite{bib58} (ICCV-23) & 74.13 & 77.13 & 70.16 & 63.98 & 69.73 & 57.03 & 63.46 & 61.98 \\
 & SADLR~\cite{bib59} (AAAI-23) & 74.24 & 76.25 & 70.06 & 64.28 & 69.09 & 55.19 & 63.60 & 63.56 \\
 & ReMember~\cite{bib11} (ECCV-24) & 74.54 & 76.74 & 70.89 & 65.00 & 70.78 & 57.53 & 63.90 & 64.00 \\
 & ASDA~\cite{bib60} (ACM MM-24) & 75.06 & 77.14 & 71.36 & \textbf{66.84} & \textbf{71.13} & 57.83 & 65.73 & 66.45 \\
 \midrule
 & \textbf{TFANet (Ours)} & \textbf{76.54} & \textbf{78.24} & \textbf{72.86} & 66.67 & 70.96 & \textbf{58.00} & \textbf{68.04} & \textbf{70.51} \\
\bottomrule
\end{tabular}
\label{table1}
\end{table*}

\section{Experiment}\label{sec4}
\subsection{Experimental Setting}\label{subsec1}
We employed VMamba as the backbone for experiments to extract visual features and utilized CLIP to extract text features. The input image size was standardized to 480×480. The maximum input length of linguistic expressions was set to 17 for RefCOCO and RefCOCO+, and 22 for G-Ref. The network was trained for 50 epochs using the AdamW optimizer, with an initial learning rate that decayed starting from the 25th epoch. All experiments were conducted on two NVIDIA RTX 4090 GPUs with a batch size of 8.

\subsection{Datasets and Metrics}\label{subsec2}
To evaluate our method, we conducted experiments on three widely used Referring Image Segmentation (RIS) datasets: RefCOCO\cite{bib13}, RefCOCO+ \cite{bib13}, and G-Ref \cite{bib14}\cite{bib15}. RefCOCO contains 19,994 images and 142,210 referring expressions, with 50,000 objects collected from the MSCOCO \cite{bib54} dataset via an interactive game interface. It has four subsets: training (120,624), validation (10,834), test A (5,657, containing persons), and test B (5,095, containing non-human objects). RefCOCO+ has similar image numbers but prohibits positional words, containing 120,191 training, 10,758 validation, 5,726 test A, and 4,889 test B samples. G-Ref, collected via Amazon Mechanical Turk, has 26,711 images, 54,822 annotated objects, and 104,560 textual descriptions, with longer and more descriptive text. We use mIoU and oIoU as primary metrics, and also employ Precision@X$(X\in\mathbb\{50,60,70,90\})$ to assess performance.

\subsection{Comparisons With the State-of-the-Art Approaches}\label{subsec3}
As summarized in Table~\ref{table1}, we compare TFANet with a set of state-of-the-art methods across three benchmark datasets. The mIoU metrics of TFANet on RefCOCO, RefCOCO+, and G-Ref show improvements of 1.84\%, 1.52\%, and 2.29\%, respectively, compared to the previous best-performing methods. This result demonstrates that the TFANet’s hierarchical alignment mechanism effectively captures both location and appearance features in RefCOCO, while also adapting to the complex and diverse linguistic expressions in RefCOCO+ and G-Ref. By incorporating multiscale attention computation, a selective long-range modeling mechanism, and linguistic information reinjection during decoding, TFANet achieves more pronounced performance gains on G-Ref datasets with greater linguistic complexity. 
To further validate and analyze the effectiveness of TFANet, we conduct three comparative experiments from different perspectives.

Compared to ReMamber (ECCV-24)\cite{bib11}, a cross-modal fusion method based on the latest Mamba framework, TFANet achieves performance improvements of 2\%, 1.67\%, and 4.14\% on the three RefCOCO-series datasets, respectively. This indicates that our progressive knowledge alignment and hierarchical feature fusion strategy outperform ReMamber's direct selective scanning mechanism. While Mamba's sequence modeling effectively captures sequential dependencies, it primarily focuses on global multimodal feature scanning and fusion, neglecting word-to-pixel correspondences, which results in suboptimal feature integration. In contrast, TFANet's multiscale attention and structured feature alignment across different granularity levels facilitate richer cross-modal interactions, ultimately enhancing segmentation performance. 

In comparison to pixel-level/region-level information modeling methods such as DDMI (ICCV-23)\cite{bib57} and ReLA (CVPR-23)\cite{bib29}, TFANet outperforms DDMI by 2.41\% in oIoU and ReLA by 4.02\% in mIoU on the RefCOCO dataset. By incorporating MLAM in the first stage, the model simultaneously captures pixel-level and region-level visual features from images, as well as multi-granularity linguistic features, including words, phrases, and sentences. This cross-modal multiscale attention computation and knowledge-plus strategy ultimately lead to significant improvements in segmentation performance. 

For the current state-of-the-art (SOTA) model ASDA, on the RefCOCO and RefCOCOg datasets, our model TFANet achieves significant performance improvements, with oIoU gains of 1.48\% and 2.31\%, respectively. This demonstrates that our TFANet effectively enhances the deep integration between vision and language. However, on the RefCOCO+ dataset, while TFANet still outperforms ASDA on the testB subset, it slightly underperforms ASDA on the validation and testA subsets. This discrepancy can be attributed to both the linguistic design of RefCOCO+ and the category composition of its subsets. Specifically, RefCOCO+ deliberately excludes all spatial expressions from its referring sentences, making models rely almost entirely on appearance-based cues such as color, size, and texture. Within this setting, the testA subset consists exclusively of human referents—such as "man," "woman," and "person"—which are commonly described using short, visually focused expressions (e.g., "the woman in red"). TFANet, with CFSM and WFDM, is particularly effective in visually complex, object-dense scenarios where deeper context modeling is required—such as those found in testB. In contrast, ASDA leverages a text-adaptive feature selection mechanism that directly enhances appearance-based visual attention, allowing it to excel in short, appearance-dominated expressions like those prevalent in testA and the validation set.

Despite this, it is worth noting that under stricter evaluation conditions—such as precision at higher IoU thresholds—ASDA exhibits a sharp decline in mask quality, as will be further discussed later. TFANet, by contrast, maintains high precision across thresholds, reflecting its advantage in generating spatially coherent and semantically aligned masks. Future research could further enhance TFANet's performance on appearance-centric subsets by integrating color-shape sensitive modules in  KFS and by developing attention mechanisms tailored for short-text linguistic structures. These refinements could improve TFANet’s responsiveness to concise, attribute-focused referring expressions while preserving its strengths in structurally ambiguous scenes.

\begin{table}[ht]
\centering
\caption{Comparison with SOTA methods on RefCOCO val}
\label{table2}
\begin{tabular*}{\columnwidth}{@{\extracolsep{\fill}}lccccc}
\toprule
Method & oIoU & mIoU & P@0.5 & P@0.7 & P@0.9 \\
\midrule
LAVT\cite{bib10} & 72.73 & 74.46 & 84.46 & 75.28 & 34.30 \\
CoupAlign\cite{bib43} & 74.70 & 75.49 & 86.40 & 77.59 & 32.40 \\
SADLR\cite{bib58} & 74.24 & 76.52 & 86.90 & 78.79 & 37.36 \\
ASDA\cite{bib60} & 75.06 & - & 86.37 & 77.74 & 29.80 \\
ReMember\cite{bib11} & 74.54 & 75.65 & 85.71 & 76.81 & 34.47 \\
TFANet & \textbf{76.54} & \textbf{77.84} & \textbf{87.59} & \textbf{81.38} & \textbf{41.29} \\
\bottomrule
\end{tabular*}
\end{table}
To comprehensively assess the effectiveness of our method, we further evaluate TFANet using multiple metrics, as shown in Table~\ref{table2}. The results demonstrate that TFANet consistently outperforms existing state-of-the-art (SOTA) models with publicly available code and weights across oIoU, mIoU, and Precision@X metrics. Specifically, we observe relative improvements of 2\% in oIoU, 2.19\% in mIoU, and 1.88\%, 4.5\%, and 6.82\% in Precision@X (X=0.5, 0.7, 0.9), respectively. Particularly, the strong performance on the strict Precision@0.9 metric highlights the effectiveness of our three-stage cross-modal interaction framework in accurately localizing linguistic referents. This performance gain stems from TFANet’s hierarchical alignment strategy, which promotes effective feature integration at multiple granularity levels. The first stage, equipped with the Multi-scale Linear Cross-Attention Module (MLAM), performs fine-grained visual-linguistic fusion, enabling rich semantic abstraction. The second stage leverages the Cross-modal Feature Scanning Module (CFSM) to model long-range dependencies, improving global contextual understanding. Finally, the Word-level Linguistic Feature-guided Semantic Deepening Module (WFDM) reinforces textual cues throughout decoding, thereby mitigating semantic degradation and enhancing mask refinement.

In contrast, although ASDA demonstrates competitive performance on coarse-grained metrics such as oIoU on the some dataset, it always falls short on stricter criteria like Precision@0.9. This indicates a limited ability to precisely localize referent objects under high-precision requirements. Such a discrepancy underscores the strength of TFANet in capturing subtle referential cues and producing high-quality segmentation masks through its progressive cross-modal alignment strategy.

\subsection{Ablation Study}\label{subsec4}
To further validate the effectiveness of TFANet, we conduct a detailed analysis of its main components, as outlined in Table~\ref{table3}. 

1)Baseline Model. As shown in Table~\ref{table3}(a), the baseline method extracts visual features from the visual encoder and fuses them with word-level linguistic features via element-wise multiplication. The fused representation is then concatenated with the original visual features and sentence-level linguistic features along the channel dimension. A convolutional layer is used to adjust the number of channels before feeding the features into a lightweight segmentation decoder to generate the final segmentation mask.

\begin{table*}[ht]
\centering
\caption{Ablation study on the RefCOCO validation, Test A and Test B datasets}
\label{table3}
\small
\renewcommand{\arraystretch}{0.7} 
\footnotesize  
\begin{tabularx}{\textwidth}{@{}lXXXXXXXXXXXX@{}}
\toprule
Method & \multicolumn{4}{c}{val} & \multicolumn{4}{c}{test A} & \multicolumn{4}{c}{test B} \\
\cmidrule(lr){2-5} \cmidrule(lr){6-9} \cmidrule(l){10-13}
 & P@0.5 & P@0.7 & P@0.9 & oIoU & P@0.5 & P@0.7 & P@0.9 & oIoU & P@0.5 & P@0.7 & P@0.9 & oIoU \\
\midrule
(a) baseline & 80.80 & 72.40 & 34.58 & 67.50 & 83.45 & 76.22 & 34.72 & 69.30 & 76.66 & 67.12 & 35.01 & 64.08 \\
(b) (a)+CFSM & 83.26 & 75.94 & 37.11 & 70.60 & 85.75 & 79.07 & 35.60 & 72.91 & 78.93 & 68.91 & 36.14 & 66.83 \\
(c) (b)+transformer & 84.37 & 77.61 & 39.13 & 71.99 & 87.47 & 81.54 & 39.28 & 75.45 & 81.02 & 72.82 & 39.86 & 69.76 \\
(d) (b)+MLAM & 85.71 & 78.82 & 40.47 & 74.65 & 88.12 & 82.15 & 39.51 & 76.81 & 81.25 & 73.19 & 40.52 & 70.97 \\
(e) (d)+WFDM(full) & \textbf{87.59} & \textbf{81.38} & \textbf{41.29} & \textbf{76.54} & \textbf{89.98} & \textbf{84.27} & \textbf{40.85} & \textbf{78.24} & \textbf{83.61} & \textbf{76.23} & \textbf{42.49} & \textbf{72.86} \\
\bottomrule
\end{tabularx}
\end{table*}

2) Impact of the Cross-modal Feature Scanning Module (CFSM). As shown in Table~\ref{table3}(b), replacing standard convolutional fusion in the baseline with CFSM, comprising CSSM and SCSM, yields oIoU gains of 3.10\%, 3.61\%, and 2.75\% on the RefCOCO val, testA, and testB sets, respectively. These improvements stem from CFSM’s ability to perform long-range, selective cross-modal fusion, enhancing feature alignment beyond local convolutional scopes. By dynamically scanning and aggregating spatial and semantic dependencies, CFSM strengthens the correspondence between linguistic cues and visual regions, resulting in more accurate and semantically aligned segmentation.

3) As shown in Table~\ref{table3}(c)(d), both standard Transformer attention and the proposed MLAM are incorporated into the baseline model (b) for enhancing visual-linguistic features prior to fusion. Results demonstrate that while Transformer-based attention yields notable gains—improving oIoU by 1.39\%, 2.54\%, and 2.93\% on the RefCOCO val, testA, and testB subsets respectively—MLAM delivers significantly superior performance. Specifically, compared to the Transformer-based variant, MLAM further improves oIoU by 2.66\%, 1.36\%, and 1.21\% on the same dataset again, respectively, achieving total gains of 4.05\%, 3.90\%, and 4.14\% over the baseline. This improvement stems from MLAM’s multi-scale bidirectional attention, which facilitates fine-grained cross-modal interaction by enriching textual features with object-level image cues and vice versa. These gains are attributed to MLAM’s bidirectional multi-scale attention computation, which not only allows finer-grained fusion of visual and textual features but also facilitates the mutual propagation of semantic cues. MLAM extracts language-guided visual attention while injecting visual priors back into linguistic representations, thereby enhancing context modeling and reducing ambiguity. 

More importantly, this is achieved with a more computationally efficient structure. MLAM achieves these accuracy improvements with notable efficiency. As shown in Table~\ref{table4}, when added to the baseline, MLAM boosts oIoU by +4.05\% while increasing inference time by only 3.8 ms and FLOPs by just 6G, maintaining practical computational cost. Compared to the Transformer variant, MLAM delivers +2.66\% higher oIoU, reduces FLOPs by 47\%, and shortens inference latency by over 6 ms, validating the efficiency of its linearized design.

\begin{table}[ht]
\centering
\caption{Efficiency analysis and comparison of MLAM on RefCOCO val}
\renewcommand{\arraystretch}{0.85} 
\label{table4}
\begin{tabular*}{\columnwidth}{@{\extracolsep{\fill}}lccccc}
\toprule
Method & oIoU & \thead{FLOPs \\ G} & \thead{Parm \\ M} & \thead{Test Time\\  
ms
} \\
\midrule
(a) baseline+CFSM & 70.60 & 201.55 &248.00 & 25.47  \\
(b) (a)+transformer & 71.99 & 389.37 &258.57 & 35.73 \\
(c) (a)+MLAM & 74.65 &  207.60& 262.98 & 29.28 \\
(d) DETRIS\cite{bib61} & 75.72 &  258.33 & 195.53 & 42.59 \\
(e) ASDA\cite{bib60} & 75.06 & 246.71 & 214.08 & 36.89 \\
(f) (c)+WFDM(full) & 76.54 &  311.30 & 268.46 & 35.87 \\
\bottomrule
\end{tabular*}
\end{table}

While Table~\ref{table4} highlights the efficiency of the proposed MLAM module, it is equally important to emphasize that TFANet as a whole also maintains strong computational efficiency in practical scenarios. Although TFANet adopts a three-stage architecture that may lead to higher aggregated parameter counts and theoretical FLOPs, each component is carefully designed with computational efficiency in mind. Unlike recent RIS models such as DETRIS and ASDA, which rely on fundation model or tightly coupled encoder-fusion mechanisms, TFANet employs a modular design that separates visual-language interaction into progressive stages. This staged alignment not only improves interpretability and modularity, but also enables more controllable computational flow and better adaptability to hardware optimization.
Despite a higher theoretical FLOPs (311.30G), TFANet achieves superior oIoU (76.54\%) and faster inference speed (35.87 ms) than DETRIS and ASDA, as summarized in Table~\ref{table4}. This highlights that TFANet’s staged, modular design not only improves segmentation accuracy, but also supports better hardware parallelism and runtime adaptability.

\begin{figure}[h]
\centering
\includegraphics[width=0.95\columnwidth]{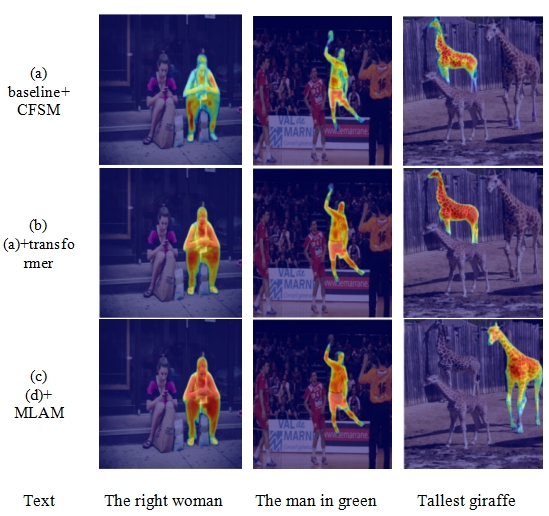} 
\caption{Visualization of attention allocation results for both the original attention mechanism and the proposed MLAM model.}
\label{fig4}
\end{figure}

\begin{figure*}[h]
\centering
\includegraphics[width=0.8\textwidth]{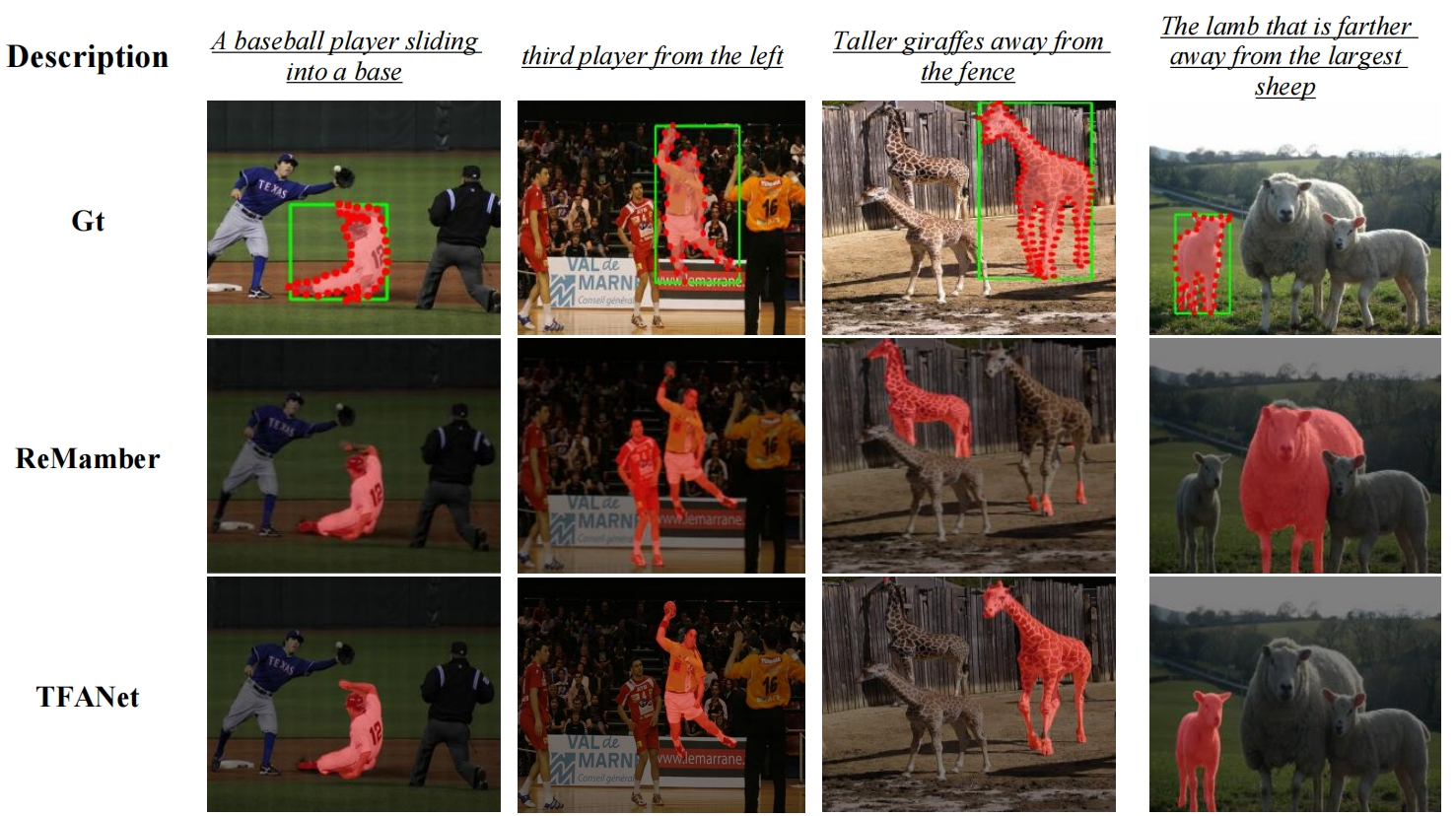} 
\caption{Visualization results of our TFANet and the baseline model ReMamber. Our model is able to predict more accurate masks.}
\label{fig5}
\end{figure*}

Moreover, as shown in Fig~\ref{fig4}, the attention allocation visualization highlights the difference between the standard attention mechanism and MLAM. The MLAM-based model exhibits greater precision in identifying MASK regions, as evidenced by a more concentrated attention distribution. This also correlates with TFANet’s higher Precision@0.9 score, demonstrating its ability to effectively localize target objects even in complex visual environments.

4) Impact of the WFDM Module. As shown in Table~\ref{table3}(e), integrating WFDM into the decoding stage yields oIoU gains of 1.89\%, 1.43\%, and 1.89\% on the three test sets over the baseline decoder in Table~\ref{table3}(d). These results demonstrate that WFDM enhances segmentation quality by leveraging word-level linguistic cues to progressively refine features. By reinforcing semantic alignment between textual descriptions and visual outputs during decoding, WFDM produces more accurate and context-aware masks, further validating its effectiveness within TFANet.

\begin{figure*}[h]
\centering
\includegraphics[width=0.8\textwidth]{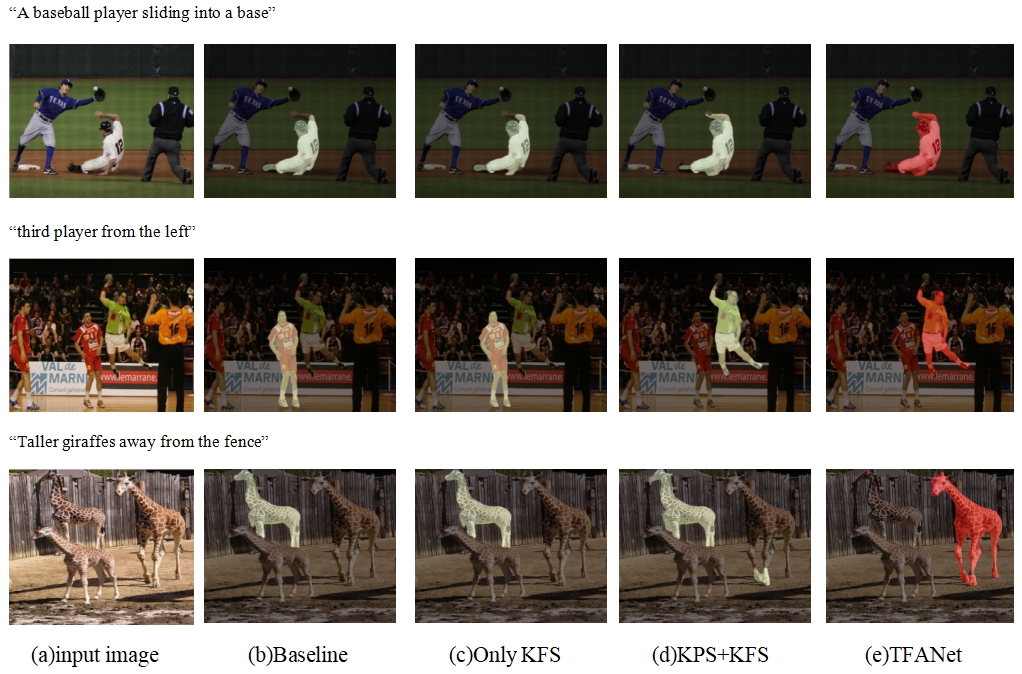} 
\caption{ Qualitative examples of the proposed components. (a) The input image and referring expression. (b) The baseline model. (c) The baseline model with KFS. (d) The baseline model with KPS+KFS. (e) The TFANet(with KPS+KFS+KIS).}
\label{fig6}
\end{figure*}

\subsection{Visualization and Qualitative Analysis}\label{subsec5}

Fig.~\ref{fig5} compares the segmentation results of our method with ReMamber\cite{bib11}, a current state-of-the-art (SOTA) model with publicly available code and pretrained weights. The visualizations show that TFANet delivers more precise segmentation, particularly in edge refinement and object boundary delineation. On RefCOCO image-text pairs, TFANet excels at capturing fine-grained details, owing to its three-phase alignment strategy (KPS, KFS, and KIS). The integration of multiscale attention (MLAM) and selective scanning (CFSM) enables stronger feature interactions, producing sharper and more accurate masks. In scenarios involving complex object relationships, TFANet further outperforms ReMamber by achieving more accurate localization and deeper contextual comprehension. This is due to its progressive cross-modal refinement, which aligns visual features with textual semantics at multiple levels. Additionally, incorporating linguistic guidance in the decoding stage restores and enriches semantic information that may be diminished during encoding. In contrast, ReMamber’s reliance on a simpler fusion strategy, primarily based on selective scanning, may lead to incomplete contextual integration and incorrect predictions, especially when multiple visually similar objects are present.

Fig~\ref{fig6} further illustrates the effectiveness of TFANet’s three-phase structure (KPS, KFS and KIS) through an ablation-based visualization study. We compare TFANet’s full model (e) with different intermediate versions, showing the progressive impact of adding key components:
(b) Baseline Model: Since the model lacks a dedicated cross-modal alignment mechanism, it frequently struggles to accurately identify target objects, often producing coarse or imprecise segmentation results.
(c) Baseline + KFS: The inclusion of KFS enhances global feature scanning, leading to improved long-range dependency modeling. This allows the model to handle more complex visual-textual interactions, but still suffers from some localization errors and coarse mask segmentation in complex situations.
(d) Baseline + KFS + KPS: Incorporating KPS introduces a multi-scale cross-attention mechanism. This mechanism enables images and text to learn a more precise allocation of attention, which facilitates finer-grained feature extraction and improved spatial-textual alignment. As a result, the segmentation masks become sharper, and the model demonstrates improved capability in distinguishing between visually similar objects.
(e) TFANet (KPS + KFS + KIS): The full model integrates all three components, with KIS complementing language knowledge and refining the segmentation boundaries through progressive linguistic feature enhancement. The masks in this setup exhibit the highest accuracy and the most complete segmentation masks, demonstrating the importance of structured hierarchical feature alignment. These qualitative visualization comparisons confirm that each additional component progressively enhances the segmentation quality, reinforcing the necessity of TFANet’s structured feature fusion strategy for accurate referring image segmentation.

\section{Conclusion}\label{sec5}
This paper proposes TFANet, a Three-Stage Image-Text Feature Alignment Network, to enhance RIS by systematically addressing cross-modal alignment inaccuracy in complex scenarios. Unlike existing approaches, TFANet refines multimodal feature alignment through a novel hierarchical three-stage learning framework improving segmentation accuracy. Specifically, the KPS incorporates a MLAM, which dynamically aligns textual knowledge from multiple linguistic and visual units with corresponding visual features while maintaining linear complexity, ensuring efficient and adaptive feature alignment. The KFS introduces a CFSM, which enhances global alignment by employing selective scanning mechanisms, effectively handling scenarios involving multiple visually similar objects. The KIS integrates a WFDM, which progressively reinforces linguistic cues throughout the mask generation process, mitigating semantic information loss and refining segmentation accuracy. Extensive experiments on RefCOCO, RefCOCO+, and G-Ref demonstrate that TFANet achieves notable gains in mIoU across all benchmarks. Ablation and visualization experiments further validate the individual contributions of each module and the overall effectiveness of the hierarchical alignment strategy. By systematically optimizing cross-modal feature alignment through a multistage learning strategy, TFANet advances RIS research by enhancing segmentation performance.

\bmhead{Acknowledgements}
This research was supported by Provincial Natural Science Foundation of Hunan grants 2023JJ30082.

\makeatletter
\makeatother

\bibliography{sn-bibliography}

\end{document}